\documentclass[runningheads]{llncs}
\usepackage[T1]{fontenc}
\usepackage{graphicx}
\usepackage{mwe} 
\usepackage{cite}
\usepackage{amssymb,amsfonts}
\usepackage{algorithmic}
\usepackage{textcomp}
\usepackage{booktabs}
\usepackage{subfigure}
\usepackage{multirow}
\usepackage{marvosym}
\usepackage{color}
\usepackage{caption}
\usepackage{newclude}
\usepackage{amsmath}
\usepackage{floatrow}
\usepackage[dvipsnames]{xcolor}
\newfloatcommand{capbtabbox}{table}[][\FBwidth]
\begin{document}
\title{Reprogramming Distillation for Medical Foundation Models}

\author{Yuhang Zhou\inst{1,2}, Siyuan Du\inst{2,3}, Haolin Li\inst{2,3}, Jiangchao Yao\inst{1,2}\textsuperscript{(\Letter)}, Ya Zhang\inst{1,2},  and Yanfeng Wang\inst{1,2}}

\authorrunning{Y. Zhou et al.}

\institute{Cooperative Medianet Innovation Center, Shanghai Jiao Tong University \and
Shanghai Artificial Intelligence Laboratory \\
\and  Fudan University\\
{\scriptsize \email{\{zhouyuhang, Sunarker, ya\_zhang, wangyanfeng622\}@sjtu.edu.cn, \\ \{dusiyuan, lihaolin\}@pjlab.org.cn} 
}}

%
%
\maketitle              
\begin{abstract}
Medical foundation models pre-trained on large-scale datasets have demonstrated powerful versatile capabilities for various tasks.
However, due to the gap between pre-training tasks (or modalities) and downstream tasks (or modalities), the real-world computation and speed constraints, it might not be straightforward to apply medical foundation models in the downstream scenarios.
Previous methods, such as parameter efficient fine-tuning (PEFT) methods and knowledge distillation (KD) methods, are unable to simultaneously address the task (or modality) inconsistency and achieve personalized lightweight deployment under diverse real-world demands.
To address the above issues, we propose a novel framework called Reprogramming Distillation (RD). 
On one hand, RD reprograms the original feature space of the foundation model so that it is more relevant to downstream scenarios, aligning tasks and modalities. 
On the other hand, through a co-training mechanism and a shared classifier, connections are established between the reprogrammed knowledge and the knowledge of student models, ensuring that the reprogrammed feature space can be smoothly mimic by the student model of different structures. 
Further, to reduce the randomness under different training conditions, we design a Centered Kernel Alignment (CKA) distillation to promote robust knowledge transfer.
Empirically, we show that on extensive datasets, RD consistently achieve superior performance compared with previous PEFT and KD methods. Source code is available at: \url{https://github.com/MediaBrain-SJTU/RD}

\end{abstract}
\section{Introduction}

Large-scale pre-training has promoted the rapid development of medical foundation models, which provides powerful features for various downstream tasks~\cite{lin2023pmc,mei2022radimagenet,nguyen2023lvm,yao2022edge,zhou2024exploring}. 
However, most medical foundation models are trained in self-supervised or weakly-supervised manners by constructing surrogate tasks like reconstruction~\cite{he2022masked}, graph modeling~\cite{nguyen2023lvm} and visual-language matching~\cite{lin2023pmc} etc. 
The resulted model cannot be directly applied to specific downstream tasks, such as tumor benign-malignant diagnosis, lung disease diagnosis, etc.
In addition, actual deployment scenarios may have more  requirements for the model's computing memory and inference speed~\cite{zhou2024low}. For example, on some edge devices or handheld instruments from different manufacturers, it is not possible to deploy large-capacity foundation models directly.
Therefore, new ways for the downstream adaptation of foundation models need to be explored.

Currently, the mainstream downstream adaptation method for foundation models is parameter efficient fine-tuning (PEFT)~\cite{xin2024parameter, chen2022adaptformer, jia2022visual, li2021prefix, hu2021lora,he2023parameter}, which typically introduces a small number of training parameters to achieve performance comparable to full fine-tuning. 
Despite its reduction in training cost, these methods might not meet the deployment constraints for memory usage and inference speed in downstream scenarios.
Knowledge distillation (KD)~\cite{hinton2015distilling,tung2019similarity,tian2019contrastive,park2019relational,yang2022masked,liu2023norm}, as an effective means of knowledge transfer, can compress the model size without sacrificing much performance of foundation models. 
However, unlike the original knowledge distillation scenario, the pre-training data used by the foundation model may be task or modality inconsistent with that of the downstream task, leading to a lack of direct correlation between  the foundation model and the student model, 
thereby weakening the gain of the foundation model for the downstream task. The close exploration~\cite{xu2023towards} through model reprogramming mainly focuses on natural scenarios, while we extend this idea into the medical scenarios with the technical improvement for an end-to-end training.

Specifically, to address the main challenge, we propose an effective framework called reprogramming distillation (RD) for adapting  medical foundation models on downstream tasks.
Specifically, on one hand, we introduce a reprogramming module to mitigate the task (or modality) inconsistency, which helps extract features more relevant to downstream tasks. 
Here, we treat the foundation model as a black box with fixed parameters, which avoids complex designs of PEFT and circumvents the significant GPU costs associated with backpropagation in the large foundation model.
On the other hand, in order to ensure that reprogrammed features can be smoothly mimic by downstream models, we adopt co-training mechanism to establish connections between the reprogrammed knowledge and the knowledge extracted by student models, and encourage them to learn similar decision boundaries through the shared classifier. 
In this process, we introduce the Centered Kernel Alignment (CKA) distillation to promote robust knowledge transfer, so as to alleviate the randomness introduced due to the various training conditions. 
In a nutshell, our contributions can be summarized:
\begin{itemize}
\item  We design a new framework called Reprogramming Distillation (RD) to facilitate downstream adaptation of foundation models. RD can align inconsistencies of tasks and modalities, achieve personalized lightweight under different deployment conditions and transfer downstream knowledge  more effectively, which cannot be achieved by ordinary PEFT and KD methods.
\item We propose two core components for RD, namely, the co-training reprogramming and the CKA distillation. The former is used to align tasks and modalities, and make the reprogrammed feature space easier mimic by downstream models; The latter is employed for robust feature transfer to reduce the randomness under different training conditions.
\item We select three different types of medical foundation models to conduct extensive experiments on five datasets with different tasks and modalities. The results indicate that RD consistently improves performance under different training conditions and significantly outperforms previous PEFT and KD methods, especially for downstream scenarios with small data size.
\end{itemize}

\section{Reprogramming distillation}

\subsection{Preliminary}
Let $F_t$ represents the medical foundation model pre-trained on broad data, namely, the teacher model, and $F_s$ represents the lightweight model to be deployed, namely, the student model.
Given the downstream data $D = \{x_i, y_i \}_{i=1}^N $, where $N$ is the training data size, 
our goal is to transfer the knowledge relevant to downstream tasks from $F_t$ to $F_s$, so that $F_s$ can not only leverage the knowledge in foundation models but also ensure efficiency in memory and computation for inference.
In the following, we present the two components of RD: the co-training reprogramming and centered kernel alignment distillation.

\subsection{Co-training Reprogramming}

In order to alleviate the inconsistency of tasks or/and modalities between pre-training data and downstream data, we propose to reprogram the feature space of the foundation model to more effectively extract knowledge relevant to downstream tasks from the foundation model. 
Basically, model reprogramming~\cite{chen2022model,xu2023towards} is a technique that involves reusing pre-trained models developed in the source domain and training only the added input transformation layers and output mapping layer to solve tasks in the target domain, thereby achieving resource-efficient cross-domain adaptation.
Inspired by this spirit, we introduce a trainable reprogramming module which includes the input transformation layers, i.e. $\phi(\cdot)$, and the output mapping layer, i.e., the classifier $g(\cdot)$, where $\phi(\cdot)$  can be any model structure and $g(\cdot)$ is a fully connected layer. Here, to ensure the simplicity and generality of the method, we use standard residual blocks for $\phi(\cdot)$.

\begin{figure}[t]
    \centering
    \includegraphics[width=0.82\linewidth]{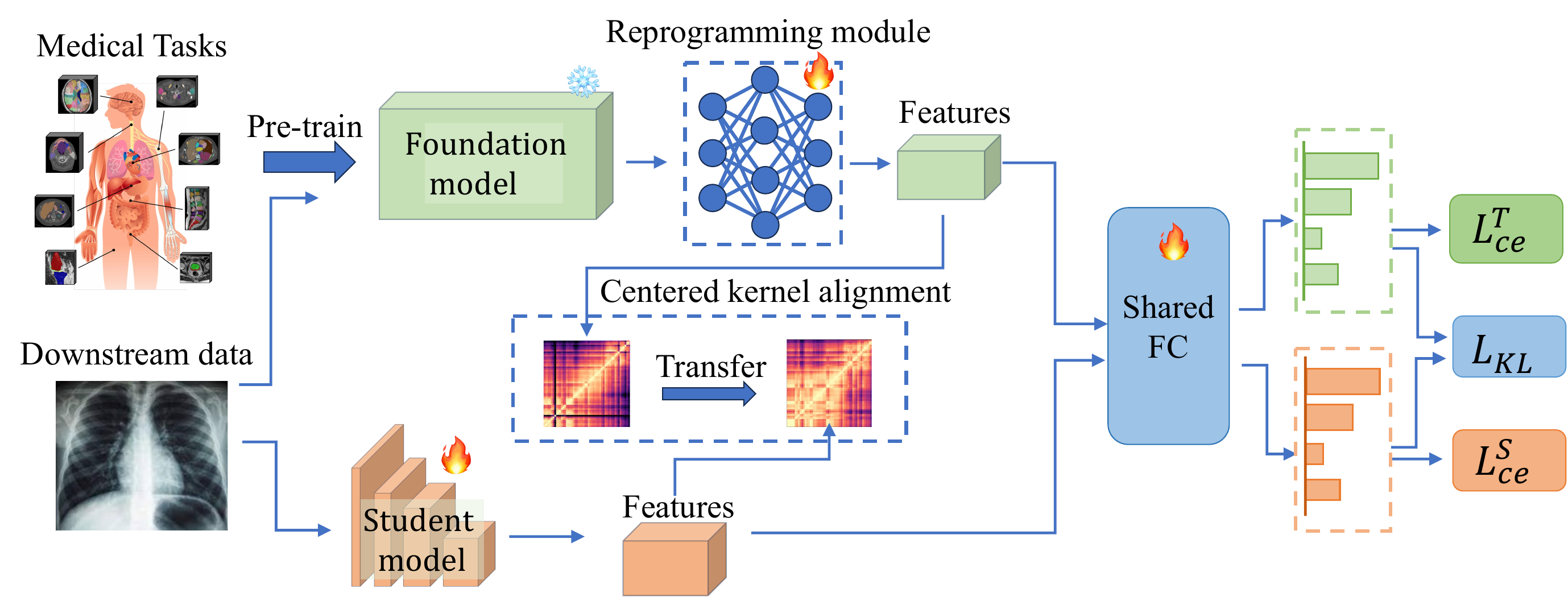}
    \caption{The overview of RD. During training, only the foundation model is fixed.}
    \label{fig:my_label}
\end{figure}

Although standard model reprogramming has been proven effective from both theoretical and experimental perspectives~\cite{chen2022model}, our goal is not only to adapt to downstream tasks but also to enable the adapted model to be deployed in a personalized and lightweight manner according to downstream medical scenarios. 
Considering that different model structures may have different tendencies in feature extraction, to ensure that the reprogrammed features can be smoothly mimic by the student model, we propose to use the co-training mechanism.
Specifically, we propose pre-training $F_s$  concurrently with training  $\phi(\cdot)$ and sharing the classifier $g(\cdot)$ to promote alignment in decision boundaries between $F_s$ and $F_t$. 
This approach acts as a conduit to enhance the relationship between the reprogrammed features and the features extracted by the student model.
The corresponding parts are illustrated in Figure~\ref{fig:my_label}.

In addition to using the standard supervised loss such as the cross-entropy loss to learn downstream tasks, we also employ logits knowledge distillation~\cite{hinton2015distilling} to enhance the relationship between the reprogrammed features and the features extracted by the student model.
Specifically, we denote the logits output by the foundation model and the student model as $z_t$ and $z_s$ respectively, then the training loss can be written as
\begin{equation}
\mathcal{L}_{train}=\mathcal{L}_{\mathrm{CE}}(y,z_s)+\alpha \mathcal{L}_{\mathrm{CE}}(y,z_t)+\beta\mathcal{L}_{\mathrm{KL}}(z_t,z_s), 
\end{equation}
where $z_s=g(F_s(x))$, $z_t=g(\phi(F_t(x)))$, $\alpha$ and $\beta$ are hyper-parameters and $\mathrm{KL}$ is the Kullback-Leibler divergence.

\subsection{Centered Kernel Alignment Distillation}

During downstream adaptation, various training conditions such as different model structures, data distributions, random seeds, etc., may introduce more unnecessary noise and increase uncertainty in feature distillation~\cite{ge2024discrepancy}. To achieve robust feature distillation, we suggest using Central Kernel Alignment (CKA) distillation.
CKA focuses on the correlation and higher-order information in features, which can reduce training instability, and is commonly used to identify corresponding layers between networks trained with different initializations~\cite{kornblith2019similarity}.

Specifically, we denote the extracted features before the shared classifier from the foundation model and the student model as $f_t$ and $f_s$ respectively, and use $K = f_t f_t^\top$ and $L = f_s f_s^\top$ to represent the pair-wise feature similarities. 
Let $\boldsymbol{H}=\boldsymbol{I}_n-\frac1n\boldsymbol{1}\boldsymbol{1}^\mathsf{T}$ be the centering matrix and $K^{\prime}= HKH$, $L^{\prime} = HLH$, where $n$ is the batch size. Then, the similarity of the centered similarity matrices can be measured by $\operatorname{HSIC}({K},{L})=\frac{{K}^{\prime}\cdot{L}^{\prime}}{(n-1)^2}$.
Here, although HSIC is invariant to orthogonal transformations of features, it is not invariant to isotropic scaling~\cite{kornblith2019similarity}. To address this, we normalize HSIC and then construct the CKA loss as follows,
\begin{equation}
\mathcal{L}_{CKA}=-\frac{\mathrm{HSIC}({K},{L})}{\sqrt{\mathrm{HSIC}({K},{K})\cdot\mathrm{HSIC}({L},{L})}}, 
\end{equation}
which can be used to determine the correspondence between hidden layers of networks trained with different random initializations and widths, while other similarity measures cannot~\cite{nguyen2020wide}.
The final training loss can be written as
\begin{equation}
\mathcal{L}_{train}=\mathcal{L}_{\mathrm{CE}}(y,z_s)+\alpha \mathcal{L}_{\mathrm{CE}}(y,z_t)+\beta(\mathcal{L}_{\mathrm{KL}}(z_t,z_s)+\mathcal{L}_{CKA}(f_t, f_s)). 
\end{equation}

\subsection{Discussion on advantages}
Compared to previous downstream adaptation methods, our RD has the following three advantages: (1) Since the trainable parameters are completely isolated from the foundation model, there is no need to backpropagate gradients through the backbone of the foundation model during training, reducing GPU usage; (2) The medical foundation model in RD can be treated as a black box, with higher parameter privacy and broader application scenarios; (3) The model obtained after downstream adaptation is lightweight, and the structure can be customized according to scenarios, making deployment more flexible. (4) It builds a CKA to promote an end-to-end training to avoid the heuristic stage-wise training~\cite{xu2023towards}.

\begin{table*}[t]
\begin{floatrow}
\capbtabbox{
\resizebox{0.48\textwidth}{!}{
\setlength{\tabcolsep}{1mm}{
\begin{tabular}{c|c|c|c|c}
\toprule
 {Dataset} &  {Task}  &{Modality} & {Classes} & {Data Size} \\ 
 \midrule
  ISIC2018  & Melanoma & RGB  &7 & 11527    \\ 
 COVID   & COVID-19    & CT  &2& 746  \\ 
 BTC   & Brain tumor   & MRI & 4 & 3264   \\ 
BUSI   & Breast cancer    & Ultrasound   &3& 626    \\ 
 ChestXray  &  Lung diseases    & X-Ray  &14 & 7135  \\ 
 \bottomrule
\end{tabular}}}
}{
 \caption{Downstream datasets with different tasks or modalities in the experiment.}
 \label{tab:pretraindata}
}
\capbtabbox{
    \resizebox{0.42\textwidth}{!}{
    \setlength{\tabcolsep}{1mm}{
    \begin{tabular}{cccc|c}
        \toprule
          \textbf{Di. Reprog.} & \textbf{Co. Reprog.} & \textbf{KD} & \textbf{CKA} & \textbf{Acc}  \\
        \midrule
          &  &  &  & 75.86 \\
        \midrule
         \checkmark &  & \checkmark &  & 80.79 \\
          & \checkmark & \checkmark &  & 82.27 \\
           & \checkmark & \checkmark & \checkmark & \textbf{85.71} \\
        \bottomrule
    \end{tabular}}}
}{
 \caption{Ablation experiments of different components in our method.}
 \label{tab:ablation_study}
 \small
}
\end{floatrow}
\end{table*}

\section{Experiments}
\subsection{Experimental Setup}
\par{\noindent \bf Foundation models.} 
We selected three medical foundation models trained in different ways to validate the proposed method's effectiveness, namely PMC-CLIP~\cite{lin2023pmc}, RadDenseNet~\cite{mei2022radimagenet}, and LVM-Med~\cite{nguyen2023lvm}. 
PMC-CLIP is trained on 1.6 million pairs of image-caption data by contrastive learning, with a visual module structure of ResNet-50. RadDenseNet is trained on 1 million images by supervised learning, with a structure of DenseNet-121. LVM-Med is trained on 1.3 million images by self-supervised learning, with a structure of VIT-B. In our experiments, the parameters of these foundation models remain fixed.

\begin{table}[th]
\caption{{Comparison with KD methods. Due to space constraints, we only display all the results on the BUSI and BTC here. The complete results for the other three datasets are presented in the form of average gains in Table \ref{tab:unlabel_ablation1}.}}
\centering
\resizebox{0.95\textwidth}{!}{
\setlength{\tabcolsep}{3mm}{
\begin{tabular}{c|c|c|cccccc|c}
\toprule
 Dataset & Student & Teacher  & VID & SemCKD & Crd & Hint & MGD & Norm & Ours  \\
\midrule
 \multirow9{*}{BUSI} &\multirow3{*}{ResNet18  77.92} & PMC-CLIP 88.31   & {77.92} & 68.83 & 75.32 & {77.92}  &77.27 &73.38& \textbf{88.31} \\
& & Densenet121 86.36   & 76.62 &70.13 & 75.97 & 77.27 &{78.57}& 68.83& \textbf{88.31} \\
 && {LVM-Med 94.16}   & 73.38 & 69.48 & 68.83 & {79.22}  & 66.23 & 64.29 & \textbf{88.96} \\
\cmidrule(r){2-10} 
 &\multirow3{*}{ShuffleNet 81.82} & {PMC-CLIP 88.31}   & {85.06} & 75.97 & 82.47 & 81.17 & 84.41 & 64.94 & \textbf{86.36} \\
 && Densenet121 86.36   & 81.82 & 76.62 & 81.17 & 83.77  & {84.41} & 56.49 & \textbf{85.71} \\
 && LVM-Med 94.16   & 81.82 & 80.52 & 79.87 & 81.17 & {82.47} & 79.87 & \textbf{85.71} \\
\cmidrule(r){2-10} 
 &\multirow3{*}{MobileNet 80.52} & PMC-CLIP 88.31   & 85.06 & 85.06 & 81.82 & 83.12 & 85.06 & 82.47 & \textbf{86.36} \\
 && {Densenet121 86.36}   & 83.12 & 84.42 & 81.17 & 83.12 & 82.46 & 61.04 & \textbf{87.01} \\
 && {LVM-Med 94.16} & 83.77   & {86.36} & 83.12 & 83.77 &  81.17 & 71.43 & \textbf{87.01} \\
\midrule
 \multirow9{*}{BTC} &\multirow3{*}{ResNet18  77.92} & PMC-CLIP 80.20   & 77.16 & 78.17 & 68.78 & 77.66  & 75.89 & 77.66 & \textbf{80.71} \\
 && Densenet121 80.46   & {77.66} & 77.41 & 76.40 & 77.16  & 76.90 & 60.41 & \textbf{78.68} \\
 && {LVM-Med 80.71} & 77.41 & 78.43 & 75.13  & {80.71} & 55.33 & 71.57 & \textbf{81.98}\\
\cmidrule(r){2-10} 
& \multirow3{*}{ShuffleNet 78.17} & PMC-CLIP 80.20   & 77.41 & 77.16 & 69.29 & 77.16  & 76.90 & 75.13 & \textbf{79.44} \\
 && Densenet121 80.46  & {77.41} & 77.16 & 76.90  & 75.63  & 76.65 & 55.08 & \textbf{79.95} \\
 && {LVM-Med 80.71}   & 76.65 & 78.68 & 75.13 & 77.92  & {79.95} & 67.77 & \textbf{80.46} \\
\cmidrule(r){2-10} 
 &\multirow3{*}{MobileNet 76.90} & PMC-CLIP 80.20   & {77.92} & 77.41 & 61.93 & 76.65  & {77.92} & 74.62 & \textbf{79.70} \\
 && Densenet121 80.46   & {78.17} & 77.41 & 76.90 & 76.90  & 77.16 & 54.82 & \textbf{79.95} \\
 && {LVM-Med 80.71}  & 77.41 & {79.44} & 74.87 & 77.92  & 72.91 & 69.04 & \textbf{82.49} \\
\midrule
  \multirow3{*}{ISIC} &\multirow3{*}{ResNet18  70.50} & {PMC-CLIP 76.92}  & 75.73 &75.66 & 76.26 & 75.33 & {77.05} &  75.79 & \textbf{78.11} \\
 && Densenet121 73.41   & 75.99 & 75.99 & 76.32 & 76.52  & {77.18} & 63.76 & \textbf{77.18} \\
 && LVM-Med 82.74  & 75.93 & 77.18 & 78.77 & {79.23} & 75.99 & 78.37 & \textbf{79.76} \\
 \midrule
   \multirow3{*}{COVID} &\multirow3{*}{ResNet18  75.86} & {PMC-CLIP 78.82}   & 77.34 & 76.85 & 74.88 & 75.37 & 75.86 & {79.80} & \textbf{81.28} \\
 && Densenet121 77.34   & 77.83 & 77.83 & 73.89 & 77.83 & {79.31} & 73.89 & \textbf{78.82} \\
 && LVM-Med 83.74  & 76.85 & {80.79} & 75.86 & 79.80 & 73.40 & {82.27} & \textbf{82.27} \\
 \midrule
   \multirow3{*}{ChestXray} &\multirow3{*}{ResNet18 92.87} & PMC-CLIP 96.11   & 92.74 & {94.16} & 93.64 & 93.13  & 93.51 & 72.24 & \textbf{95.33} \\
 && Densenet121 94.55   & 93.77 & 92.35 & {94.03} & 93.39  & 92.09 & 90.53 & \textbf{94.68} \\
 && {LVM-Med 94.16}    & 93.64 & 93.51 & 93.51  & {94.42} & 92.22 & 81.84 & \textbf{96.11} \\
\bottomrule
\end{tabular}}}
\label{table_kd}
\end{table}

\begin{table}[th]
\caption{Comparison with PEFT methods. Since PEFT is mainly applied to transformer structures, the foundation model in this experiment is LVM-Med.}
\centering
\resizebox{0.85\textwidth}{!}{
\setlength{\tabcolsep}{4mm}{
\begin{tabular}{c|c|ccccc|c}
\toprule
 Dataset & Student & AdaptFormer & MeLo & VPT & VQT & LPT & Ours  \\
\midrule
 \multirow3{*}{BUSI} & ResNet18 & 80.20 & 86.36 & 86.36 & 87.66 & 87.66 & \textbf{88.96}\\
  & ShuffleNet & 78.93 & 83.77 & 84.77 & 84.77 & 84.42 & \textbf{85.71}\\
  & MobileNet & 79.95 & 84.42 & 81.82 & 85.06 & 81.82 & \textbf{87.01}\\
  \cmidrule(r){1-8} 
  \multirow3{*}{ISIC} & ResNet18 & \textbf{79.76} & 76.65 & 77.84 & 77.12 & 76.79 & \textbf{79.76}\\
  & ShuffleNet & \textbf{78.90} &  76.19 & 77.05 & 77.05 & 76.26 & \textbf{78.90}\\
  & MobileNet & 76.79 &  \textbf{76.98} & 76.52 & 76.46 & 76.92 & {76.79}\\
  \cmidrule(r){1-8} 
  \multirow3{*}{COVID} & ResNet18 & 80.30 & 75.86 & 76.35 & 76.35 & 80.79 & \textbf{82.27}\\
  & ShuffleNet & 81.28 & 76.35 & 78.82 & 78.82 & 81.77 & \textbf{85.22}\\
  & MobileNet & 78.82 & 77.34 & 77.34 & 79.80 & 75.85 & \textbf{84.24}\\
    \cmidrule(r){1-8} 
  \multirow3{*}{BTC} & ResNet18 & 80.20 & 78.43 & 77.92 & 77.92 & 78.93 & \textbf{81.98}\\
  & ShuffleNet & 78.93 & 76.90 & 77.66 & 76.90 & 78.43 & \textbf{80.46}\\
  & MobileNet & 79.95 & 76.65 & 76.65 & 78.93 & 76.40 & \textbf{82.49}\\
\cmidrule(r){1-8} 
  \multicolumn{2}{c|}{Average} & 79.50 & 78.83 & 79.09 & 79.74 & 79.67 & \textbf{82.82}\\
\bottomrule
\end{tabular}}}
\label{table_peft}
\end{table}

\par{\noindent \bf Downstream Datasets and Lightweight Student Models.} 
To comprehensively validate the generalization of our method, we selected five medical image datasets with various modalities and tasks as the targets for downstream adaptation, namely BUSI~\cite{al2020busi}, ISIC~\cite{tschandl2018ISIC}, Covid~\cite{xingyi2020covid_CT}, BTC~\cite{saleh2020BTC} and ChestXray~\cite{cohen2020covid_xray}, which are summarized in Table \ref{tab:pretraindata}.
In the absence of official test sets, we divide the data as training and test sets in a 4:1 ratio. For the downstream  models, we use three lightweight structures, namely ResNet18, MobileNet, and ShuffleNet.

\par{\noindent \bf Implementation details.} 
We use AdamW optimizer with the learning rate set to 5e-3. Each downstream dateset is trained for 240 epochs and the batch size is set to 32. The weights $\alpha$ and $\beta$ in the loss function are initially 1 and decreases linearly with the training process. The GPUs we used are RTX 3090.

\subsection{Ablation Studies}
We verify the effectiveness of the different components in Table~\ref{tab:ablation_study}, where ``Di. Reprog'' and ``Co. Reprog'' represent direct reprogramming~\cite{chen2022model} and our co-training reprogramming respectively. 
It can be seen that replacing direct reprogramming with our co-training reprogramming results in a significant performance improvement. This indicates that the design of co-training reprogramming can better transfer knowledge to downstream models while mitigating task and modality inconsistencies. The CKA distillation also significantly improves performance, which may be attributed to robust knowledge transfer that tends to focus on the effective components in features, reducing the impact of randomness on training.

\begin{table*}[t]
\begin{floatrow}
\capbtabbox{
\resizebox{0.48\textwidth}{!}{
    \setlength{\tabcolsep}{1mm}{
    \begin{tabular}{c|c|c|c|c|c}
    \toprule
     Dataset & BUSI  & COVID  & BTC & ChestXray  & ISIC\\
    \midrule
     Data Size & 626  & 746  & 3264 & 6326 & 11527 \\
    \midrule
     \small{PMC-CLIP} &  \textbf{6.71} & 4.93 & 2.11 & \textbf{2.42} & 5.80\\
     \small{RadDensenet} &  5.19 & 6.08 & 1.86 & 0.77 & 5.10\\
     \small{LVM-Med} &  6.06 & \textbf{6.57} & \textbf{3.98} & \textbf{2.42} & \textbf{6.02}\\
    \bottomrule
    \end{tabular}}}
}{
 \caption{The average gain of three foundation models on downstream datasets. }
 \label{tab:unlabel_ablation1}
}
\capbtabbox{
\resizebox{0.42\textwidth}{!}{
    \setlength{\tabcolsep}{1mm}{
    \begin{tabular}{c|c|c|c}
    \toprule
     Cost & Params(M)  & FLOPs(G)  & Time(ms)\\
    \midrule
     \small{ \textcolor{brown}{LVM-Med}} &  41.26  & 65.34 & 77.92\\
     \midrule
     \small{ \textcolor{purple}{ResNet18}} &  5.33  & 6.77 & 3.13\\
     \small{ \textcolor{purple}{ShuffleNet}} &  0.60  & 0.55 & 7.59\\
     \small{ \textcolor{purple}{MobileNet}} &  1.06  & 1.17 & 5.95\\
    \bottomrule
    \end{tabular}
    }}
}{
 \caption{Comparison on computation and speed of models using \textcolor{purple}{RD} or \textcolor{brown}{not}. }
 \label{tab:unlabel_ablation2}
 \small
}
\end{floatrow}
\end{table*}

\subsection{Comparisons on Downstream Tasks}
\par{\noindent \bf Compared with KD methods.} 
We compare the current representative KD methods, including Hint~\cite{romero2014fitnets}, AT~\cite{zagoruyko2016paying}, VID~\cite{ahn2019variational},  SemCKD~\cite{chen2021cross}, Crd~\cite{tian2019contrastive},  MGD~\cite{yang2022masked} and Norm~\cite{liu2023norm} in the Table~\ref{table_kd}.
In addition, we also report the full fine-tuning performance on the downstream dataset directly using the student model and the foundation model behind the name of model structures in the Table~\ref{table_kd}.

It can be seen that RD is superior overall to previous KD methods, even approaching the performance of fully fine-tuning foundation models, with more lightweight structures. Compared to directly training the student model, KD methods have limited improvement and obvious differences in performance gains on different datasets.
This indicates that the direct knowledge distillation without eliminating inconsistencies of tasks or modalities is low-efficient, since the effectiveness of knowledge transfer may directly depend on the matching degree between the feature space of the foundation model and downstream tasks. 
The consistent improvement of RD on different datasets and structures demonstrates that RD is effective for downstream adaptation of the foundation model.

\par{\noindent \bf Compared with PEFT methods.} 
We compare RD with PEFT methods of different types, such as adapter tuning method AdaptFormer~\cite{chen2022adaptformer}, LoRA method MeLo~\cite{zhu2023melo}, prompt tuning methods VPT~\cite{jia2022visual} and LPT~\cite{dong2022lpt}, and prefix tuning method VQT~\cite{tu2023visual}  in the Table~\ref{table_peft}. 
For a fair comparison, these methods also employ knowledge distillation to obtain the same student models.
It can be seen that the performance of these PEFT methods varies insignificantly, within 1\%, while RD demonstrates a very noticeable performance advantage, especially for small datasets like BUSI and COVID. As for ISIC, PEFT methods and RD are comparable in performance, indicating that the effectiveness of PEFT methods relies on a sufficient amount of training data. Our method reduces the reliance on data, which is of significant importance for data-expensive medical scenarios.

\begin{figure}[t]
    \centering
    \caption{Comparison of decision boundaries of different methods. Our method has decision boundaries that are more similar to the foundation model compared to other methods, which could be a reason for the better performance of ours.}
    \includegraphics[width=0.72\linewidth]{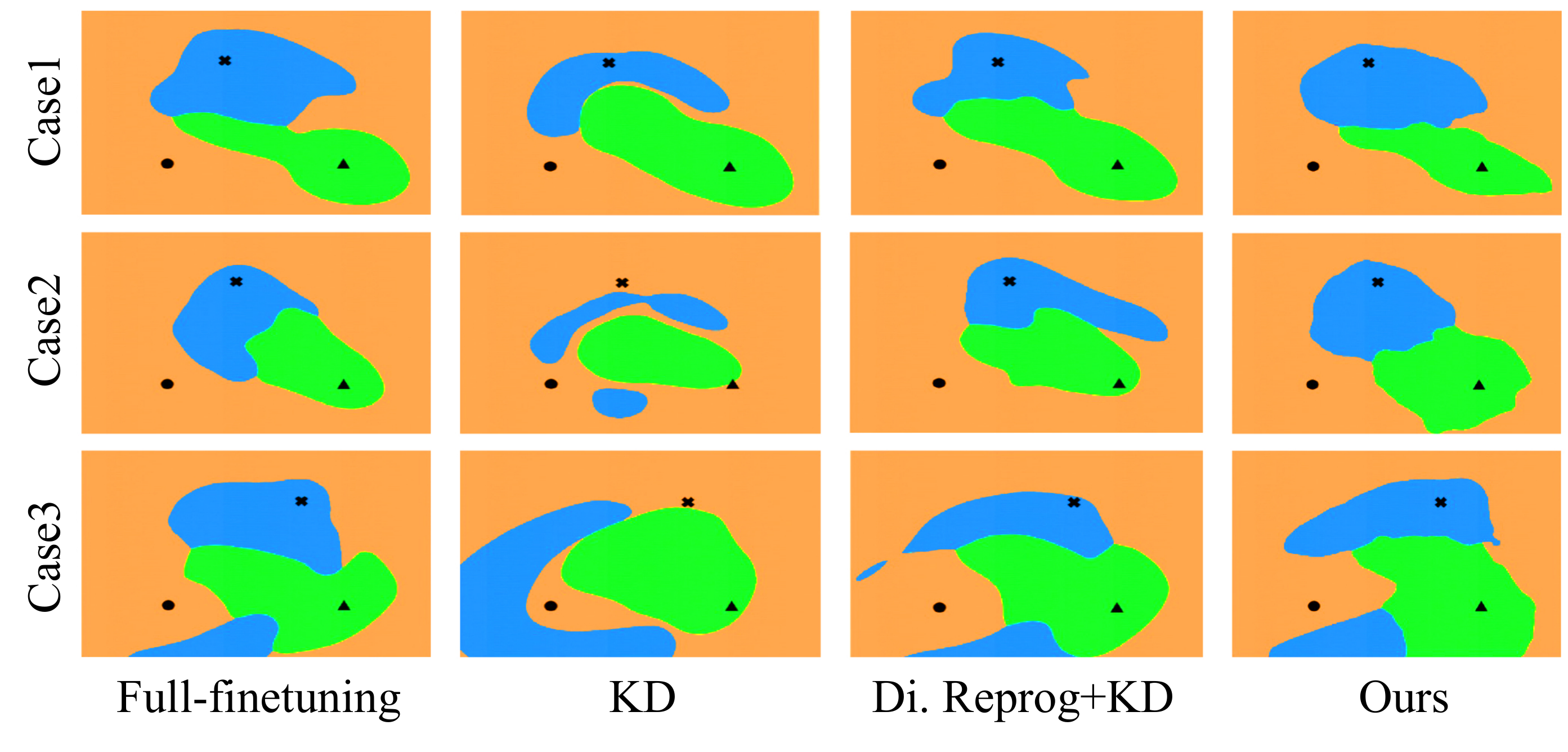}
    \label{fig:decision}
\end{figure}

\subsection{Further analysis}

Due to limited space, the complete results of RD on five downstream datasets, three foundation models, and three downstream models is calculated and summarized in  Table~\ref{tab:unlabel_ablation1}. It can be seen that RD shows the most significant performance improvement when combined with PMC-CLIP or LVM-Med. 
This indicates that the gains for downstream tasks from the foundation model obtained through self-supervised learning (SSL) are greater, demonstrating stronger generalization capability of SSL. 
Furthermore, the performance improvement is more pronounced in downstream tasks with small data size, such as BUSI, indicating that RD can effectively reduce the demand of downstream training data.

Table~\ref{tab:unlabel_ablation2} shows the cost comparison of deployed models  using RD or not, where ``Time'' represents the time taken to inference each minibatch data.
Combining it with Table~\ref{table_kd}, it can be seen that RD can achieve performance comparable to full-finetuning the foundation model and significantly reduce the computational cost during inference, improving the efficiency of the deployed model.

We also compared the decision boundaries of the trained models following~\cite{somepalli2022can}, as shown in Figure~\ref{fig:decision}. 
It can be seen that ours has decision boundaries more similar to those of fully fine-tuned foundation models compared to other baselines. 
This indicates that RD can better transfer knowledge from foundation models, achieving similar training effects as full-finetuning in a lightweight form.

\section{Conclusion}
To construct efficient and computation-friendly adaptation of the medical foundation model on the downstream tasks, it is necessary to consider the knowledge transfer under the inconsistency of data distribution and model structure. 
Different from previous PEFT methods and KD methods that cannot simultaneously address the task (or modality) inconsistency and achieve personalized lightweight deployment according to different real-world demands, we propose a novel framework RD for foundation models, which can break the constraints of data distributions and model structures. Empirically, a large number of experiments on five different modal downstream tasks, three different types of models demonstrate the generalization and effectiveness of our method.

\subsubsection{Acknowledgement.}
This work is supported by the National Key R\&D Program of China (No. 2022ZD0160703),  STCSM (No. 22511106101, No. 22511105700, No. 21DZ1100100), 111 plan (No. BP0719010) and National Natural Science Foundation of China (No. 62306178).

\subsubsection{\discintname}
The authors have no competing interests to declare that are
relevant to the content of this article.

\bibliographystyle{splncs04}
\bibliography{Paper-0912}

\end{document}